\documentclass[10pt,twocolumn,letterpaper]{article}

\usepackage{ijcb}
\usepackage{times}
\usepackage{epsfig}
\usepackage{graphicx}
\usepackage{amsmath}
\usepackage{amssymb}
\usepackage{enumitem}
\usepackage[pagebackref]{hyperref}
\usepackage[disable]{todonotes} 
\usepackage{comment}
\usepackage[accsupp]{axessibility} 

\ijcbfinalcopy 

\ifijcbfinal\fi
\begin{document}

\title{From Data to Insights: A Covariate Analysis of the IARPA BRIAR Dataset for Multimodal Biometric Recognition Algorithms at Altitude and Range}

\author{David S. Bolme, Deniz Aykac, Ryan Shivers, 
Joel Brogan, Nell Barber, \\
Bob Zhang, Laura Davies, and 
David Cornett III \\ 
Oak Ridge National Laboratory \\
1 Bethal Valley Road, Oak Ridge, TN 37830, USA \\
\texttt{bolmeds@ornl.gov}
}

\maketitle
\thispagestyle{empty}

\begin{abstract}

This paper examines covariate effects on fused whole body biometrics performance in the IARPA BRIAR dataset, specifically focusing on UAV platforms, elevated positions, and distances up to 1000 meters. The dataset includes outdoor videos compared with indoor images and controlled gait recordings. Normalized raw fusion scores relate directly to predicted false accept rates (FAR), offering an intuitive means for interpreting model results. A linear model is developed to predict biometric algorithm scores, analyzing their performance to identify the most influential covariates on accuracy at altitude and range. Weather factors like temperature, wind speed, solar loading, and turbulence are also investigated in this analysis. The study found that resolution and camera distance best predicted accuracy and findings can guide future research and development efforts in long-range/elevated/UAV biometrics and support the creation of more reliable and robust systems for national security and other critical domains.

\end{abstract}

\listoftodos

\todo[inline]{Due to IJCB: 7/3   POC:    Submission Process: https://sites.google.com/kitware.com/ijcb-lrr-session-2024/home}

\section{Introduction}

This paper investigates covariate analysis for the IARPA BRIAR program, which has crucial implications for intelligence operations such as counterterrorism, infrastructure protection, military force defense, and border security. The BRIAR program aims to develop biometric systems that can overcome image quality challenges, accommodate a broad range of yaw and pitch angle viewpoints on individuals, and integrate information from multiple sources (face, body, and gait features) without relying solely on one modality. While Nalty et al.~\cite{nalty2022brief} provide a survey of relevant techniques, this paper concentrates on analyzing algorithm performance to identify factors influencing accuracy and inform future enhancements in critical domains.

\begin{figure}[t]
    \centering
    \includegraphics[width=1.0\linewidth]{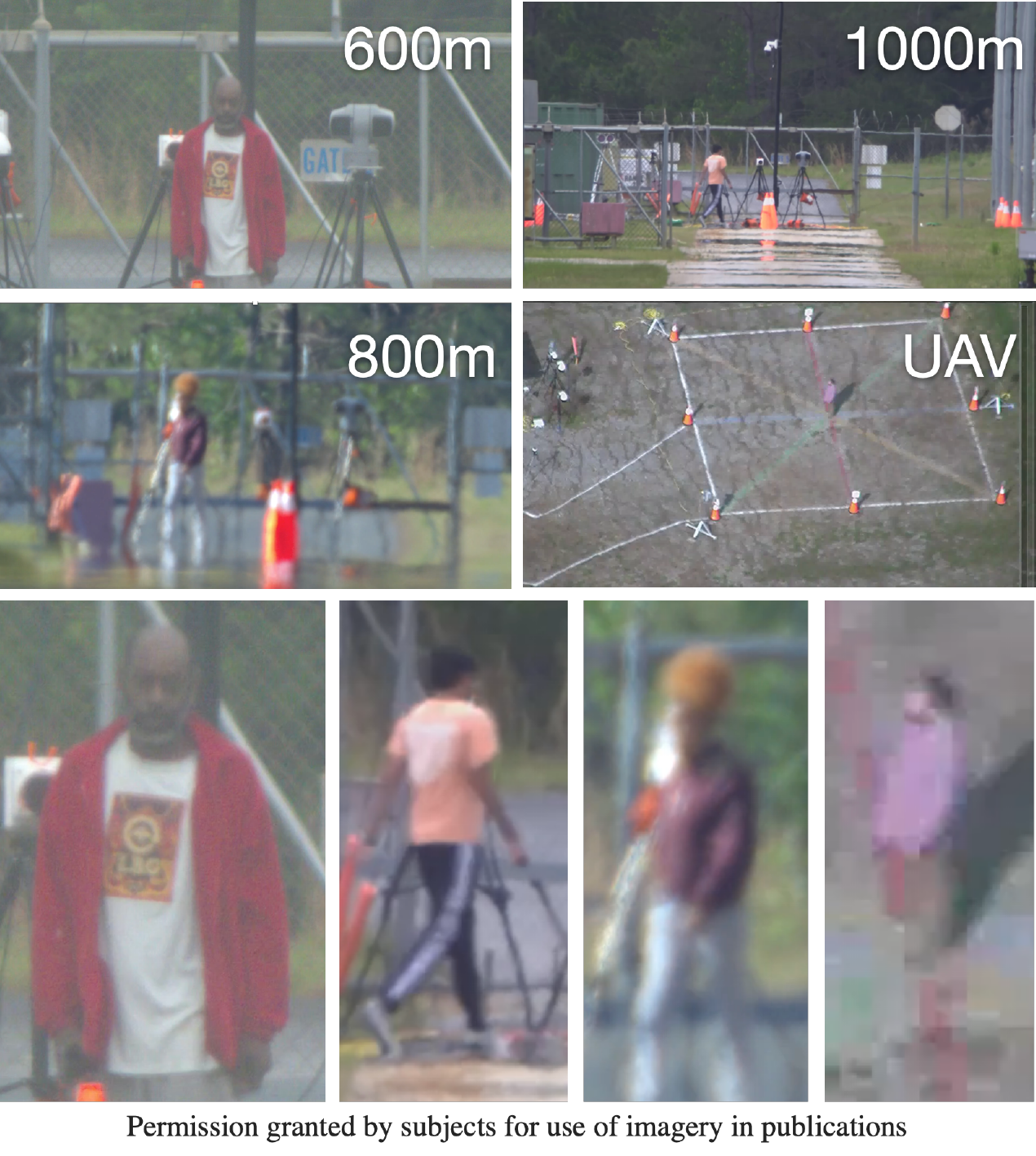}
    \caption{This figure illustrates the varying image quality in videos from the second dataset. At 600~m and 1000~m, individuals have distinct features with minimal distortion. However, at 800~m, some turbulence-induced degradation becomes apparent. The UAV captures present numerous quality issues, such as low resolution, atmospheric disturbances, and compression artifacts.}
    \label{fig:imagery-overview}
\end{figure}

This research examines five biometric matching algorithms from the BRIAR program. Although the internal operations of these systems remain proprietary, developers have shared some information through scientific publications or open-source software \cite{nalty2022brief, liu2024farsight, Zhu_2024_WACV, nikhal2023weakly, 10449024, myers2023recognizing, huang2023whole, Azad_2024_CVPR, doers2023}. For this analysis, we treat these systems as black boxes, not exploring their specific algorithms. Our study aims to gain insight into their underlying performance by analyzing system outputs and applying statistical techniques.

This study offers in-depth insights into the performance of fused whole body biometric recognition systems by analyzing covariates such as elevation, ground conditions, resolution, camera configuration, range, and various atmospheric factors like temperature, wind speed, solar loading, and turbulence. The presented results can guide future research and development efforts in biometrics while informing decision-making processes for practitioners and policymakers in national security and critical infrastructure protection. In this analysis, we discovered that the resolution, measured as the height of the head in pixels, and the distance between the camera and the subject had the most correlation with recognition accuracy. Other factors related to weather and turbulence had less influence on accuracy. 

This work presents a novel and timely exploration of the challenges and possibilities in complex and dynamic environments for biometric identification. Employing distinctive evaluation methods, Generalized Linear Mixed Models (GLMM) are applied to biometric algorithm performance, while this analysis addresses new sets of intricate problems with confounding factors. Untangling these covariate correlations is crucial in comprehending algorithm performance and the techniques utilized herein are likely applicable to future biometrics research. Additionally, score normalization methods implemented prior to modeling produce results within an easily comprehensible space.

\section{Dataset Composition}

The data used for this analysis was obtained using the techniques described in \cite{cornett2023expanding}. At the time of Cornett et al.'s publication they had collected data at two locations, but this analysis includes two additional collections, expanding the datasets used to four locations. The methods used for collecting this additional data are similar to those employed initially. The standardized and consistent nature of these collections provides a strong basis for developing a linear model to predict biometric recognition algorithm scores. The inclusion of multiple gait sequences enhances the datasets' usefulness, as gait is a challenging modality for biometric recognition systems, and its analysis can offer valuable insights into algorithm performance in this area.

The datasets include a wide range of biometric data collected in various settings across the US and at different times of the year to capture a broad spectrum of environmental and climatic conditions. These datasets consist of outdoor videos and are compared to indoor mugshot and controlled gait data. The four datasets used in this study represent approximately half of the total datasets planned to be collected by the program, providing a comprehensive and diverse sample for analysis. Each collection consists of approximately 450 individuals, offering a substantial sample size for analysis. About 200 individuals are utilized for training algorithms. For evaluation, 250 other individuals are used, while 100 are reserved for probes, which include the difficult outdoor and long-range data. An additional 150 individuals are included in the gallery consisting of only indoor data which boosts the size of the gallery.

The controlled indoor collections of face, body photos, and multiple gait sequences are used for enrollment in the gallery. This high-quality and consistent data is used for matching purposes. Each person in the gallery has facial images from five different angles, high-resolution whole-body photos from eight angles, and gait videos from a variety of perspectives and elevations. While there may be some cases with missing images or videos, the algorithms are given all available information to generate a single entry per person in the gallery database containing multiple biometric signatures and embeddings to support the modalities provided. 

The outdoor data collections, used for probes, capture subjects participating in various activities to focus on the inherent biometric signatures of individuals. These data collections take place in a 10-meter square box equipped with multiple camera systems and allow for a comprehensive capture of subjects' appearances. Each probe is a 5- to 15-second video, typically capturing a portion of an activity. The subjects are instructed to perform a range of activities within the box, including standing, walking, using a cellphone, moving boxes, and other daily actions. Capturing subjects' movements and behaviors in an outdoor environment provides a wealth of information for analyzing covariates and their impact on biometric recognition performance.

\subsection{Experimental Protocol Composition}

The BRIAR dataset is still expanding and the analysis presented here is based on the BRIAR experimental protocol version 4.2.1, which includes  9,215 clips featuring 371 subjects. To conduct the analysis, 5- to 15-second video clips are extracted from the captured activities and matched to a gallery of indoor data collected in controlled environments. To account for larger searches and enhance the accuracy of the analysis, the gallery is supplemented with an additional 487 subjects, providing a more substantial basis for the evaluation of algorithm performance. The gallery type used in the analysis is called ``simple'' which consists of face and whole body images and videos of subjects walking in a predefined straight path. By comparing the extracted video clips to the indoor gallery, the analysis will reveal valuable insights into the impact of covariates on biometric recognition performance, particularly in relation to gait and behavioral patterns. This comprehensive and systematic approach will pave the way for a more robust and reliable development of biometric recognition systems, particularly in complex and dynamic environments.


\section{Score Normalization}

Normalization is a critical step in preparing the data for the analysis of covariates in the IARPA BRIAR dataset. The purpose of normalization is two-fold. Firstly, it allows us to transform the scores generated by the five experimental biometric recognition algorithms into a common scale, making them comparable and ensuring that they are in the same compatible score space. This is essential for a fair and accurate analysis of algorithm performance, as the raw scores may vary significantly from one algorithm to another due to different underlying processing mechanisms.

Secondly, the normalization scheme also provides valuable insights and intuition into interpreting the results of the linear model. The model will predict the effect of the covariates on the genuine distribution of the biometric match scores, while it is assumed that the impostor distribution is essentially stable. Because these covariates relate to quality changes in the biometric data, they will shift the match distribution in one direction or the other. It is easiest to interpret this as a shift in the receiver operating characteristic (ROC) curve: If the quality of the image goes down, the ROC curve will also shift down.

Since a given score determines a single point on that ROC curve, the model will essentially be predicting how the ROC curve shifts as covariate values change. Because the statistical models used will be predicting the "expected genuine score" for a given set of covariates, we can interpret model results as predicting a single point of the resulting ROC curve. As the statistical model is predicting the center point of the genuine distribution, the predicted (True Accept Rate) TAR will be 50\% because approximately half of the distribution should fall above and below that predicted score. Therefore, the TAR is essentially fixed.

Determining the false accept rate (FAR) is where normalization comes in. The normalization selected converts the scores into the log of the estimated FAR ($\log_{10}(\text{FAR})$) so that the predicted score directly correlates to FAR and can be used to interpret the effects on the ROC curve.

Normalization involves a series of steps aimed at transforming the raw scores generated by the biometric recognition algorithms into a standardized format. The scores are typically associated with verification results and, in particular, the expected FAR and TAR values. These metrics are computed for the verification performance metric over the evaluation dataset.

\begin{figure*}
    \centering
    \includegraphics[width=0.75\linewidth]{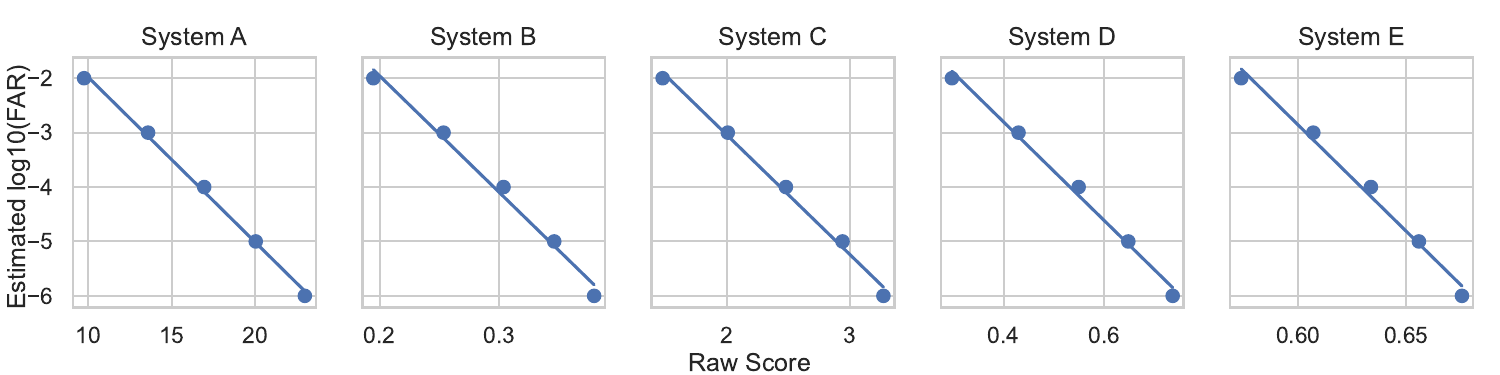}
    \includegraphics[width=0.75\linewidth]{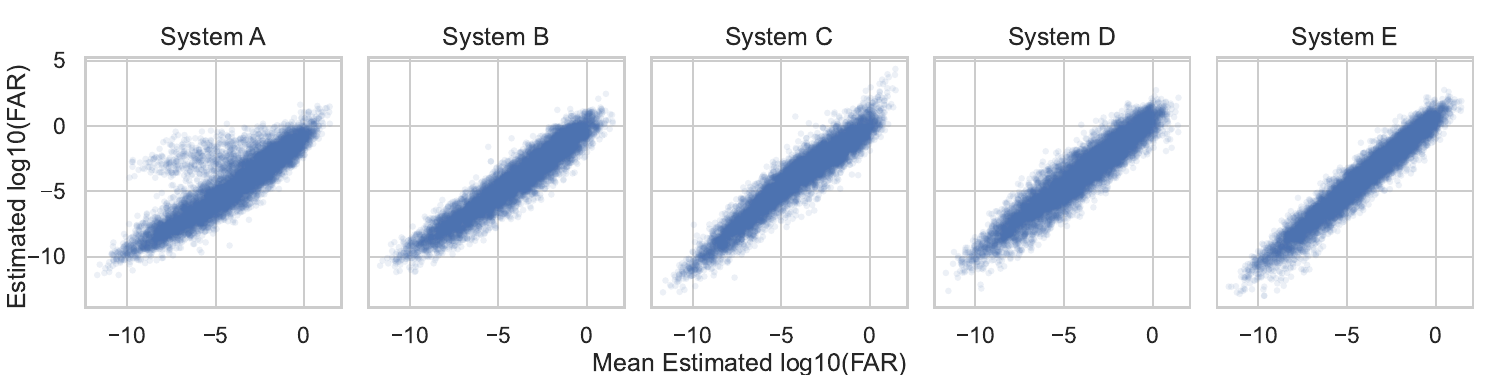}
    \includegraphics[width=0.75\linewidth]{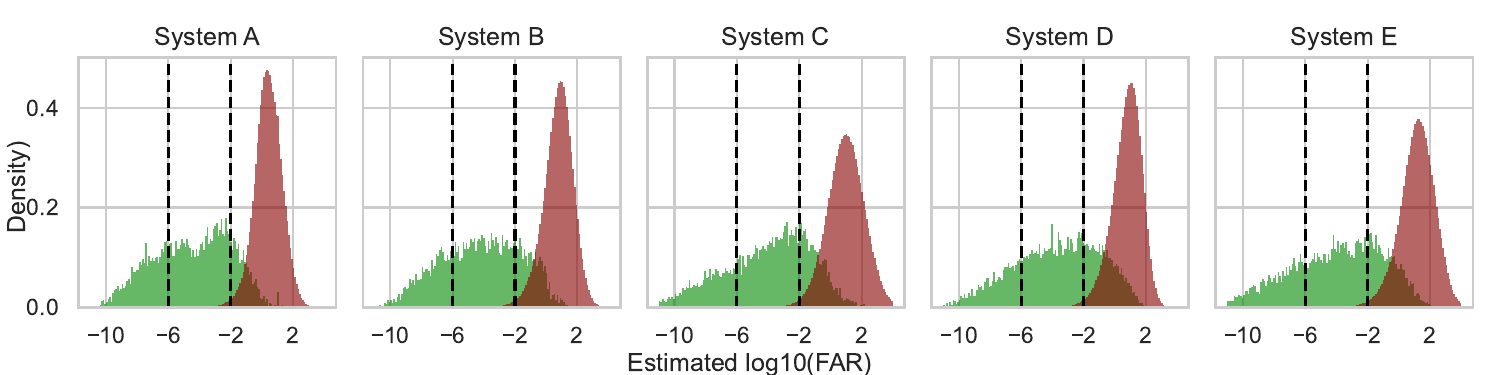}
    \caption{Top: This row shows that the score to $\log_{10}(\text{FAR})$ fits are linear for all systems.  Middle: This row shows that all systems scores correlate strongly with the mean with a potential exception of System A which shows some unique behavior. Bottom: This row shows the genuine and impostor score distributions after normalization with the region used for normalization within the dotted lines.}
    \label{fig:score-normalization}
\end{figure*}

The normalization process focuses on the estimates of FAR in the tail of the impostor distribution. The FAR represents the probability of the system incorrectly accepting an impostor as a genuine person. To standardize the FAR values we fit the following linear transformation, where $m$ and $b$ are least squared line fits computed for each algorithm at five estimate of FAR $10^{-6}$, $10^{-5}$, $10^{-4}$, $10^{-3}$, and $10^{-2}$.  This region of the impostor distribution (the red distributions in Figure \ref{fig:score-normalization}) represents 1\% of the impostor scores where the PDF approximately follows a exponential curve and can therefore be represented following model.

\begin{equation}
\log_{10}(\text{FAR}) = m \times \text{score} + b
\end{equation}

The normalization process is primarily focused on the FAR values in the tail of the impostor distribution, specifically the region where $FAR < 10^{-2}$. This region is particularly important for several reasons. Firstly, it is in this range that most operational applications are interested in, as they typically require very small FAR values. Secondly, in this range, the relationship between $\log_{10}(FAR)$ and the raw scores from the algorithm is approximately linear, as demonstrated in Figure~\ref{fig:score-normalization}.

This linear relationship is significant because it enables us to analyze and predict the expected genuine scores for a given set of covariate values by applying this normalization. It should be noted, however, that scores larger than $10^{-2}$ will no longer have a linear fit (although this region is not of interest), and scores below $10^{-6}$ will be extrapolated, potentially resulting in less accurate predictions for very small FAR values. 

\section{Correlated Variables and Interactions}
\label{sec:correlated_variables}

Before investigating the effects of other covariates, it is crucial to examine the specific influences of the sensor model and collection location on the algorithm's performance. The data collection procedures and environmental factors as part of the datasets may introduce several quality issues and interactions that need to be taken into account when analyzing covariates.

Many of the variables and metadata collected are correlated, which makes the analysis more challenging. To ensure the validity and reliability of the findings, it is essential to account for confounding variables. This will help in understanding the true impact of the \textbf{sensor model} and \textbf{collection location} on the algorithm's performance while considering the potential influence of other factors. By doing so, a more comprehensive assessment of the algorithm's performance can be obtained.

\begin{figure}
    \centering
    \includegraphics[width=1\linewidth]{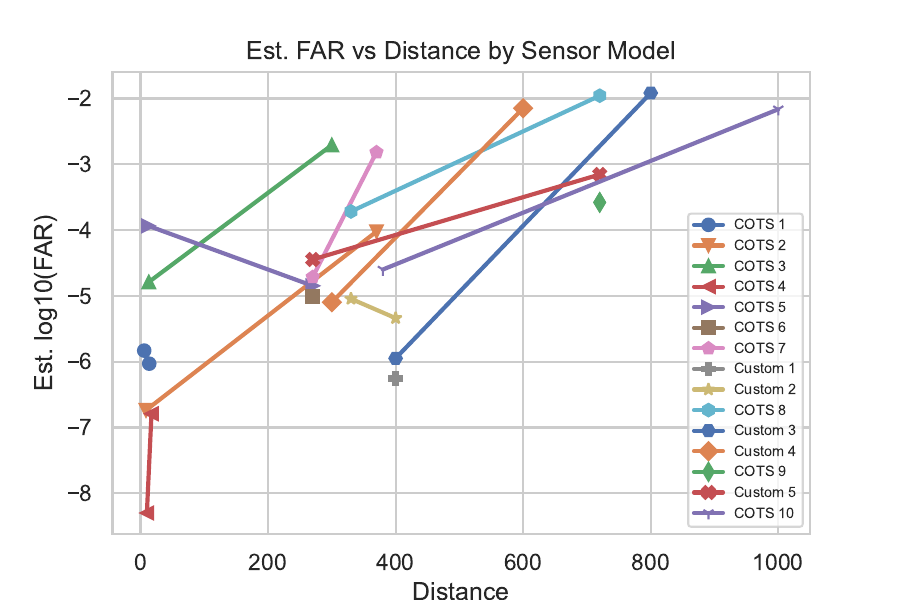}
    \caption{This figure shows the interactions between sensor model and distance.  Also shown is the distance range at which each sensor has been deployed.}
    \label{fig:sensor-distance}
\end{figure}

The selection and configuration of sensors play a crucial role in determining accuracy. As illustrated in Figure \ref{fig:sensor-distance}, a simple model demonstrates how the sensor's design interacts with distance. Specifically, the size and configuration of a sensor's optics greatly influence the distance at which it can effectively operate, leading to a strong correlation with distance. Additionally, there is a wide variation in accuracy, which is primarily due to differences in sensor quality or how the sensor is configured.

The value of analyzing sensor models cannot be overstated when it comes to organizations aiming to develop new camera systems. Careful selection of cameras can have a significant impact on overall performance. Existing systems can be updated or reconfigured to enhance biometric performance. When procuring new systems, it is important to note that the specific models tested here may no longer be available as these products are constantly being updated and improved. However, accurately modeling other covariates necessitates a meticulous process of isolating and understanding the correlation and quality variations in sensors. By thoroughly examining these factors, organizations can optimize their systems and achieve superior performance.

The dataset includes a significant influencing factor in the form of the location where the test data was gathered. Data is collected at various sites and throughout different times of the year, with each collection correlated with season, weather, climate, geography, and local demographics. At each site, cameras are repositioned to suit the collection location, resulting in a correlation with camera distance and configuration. Each collection site also imposes unique constraints on the types of UAVs that can be flown and their operational parameters.

Consequently, the collection location must also be taken into account when assessing algorithm performance. Unlike camera selection and configuration, the collection location, weather, and other factors cannot be easily controlled. Thus, operational systems must be capable of functioning in all conditions.

To effectively model algorithm performance, it is crucial to consider the measured weather data and other general parameters while minimizing the impact of the specific location and time of year where the data was collected. By doing so, more accurate and reliable predictions can be made.

We recognize that each challenge presented by the IARPA BRIAR dataset - for example, close-range security checks, long-range recognition, elevated cameras, and moving UAV platforms - is worthy of its own research program. The analysis of covariates in the IARPA BRIAR dataset presents a challenging and complex task, requiring  deep understanding of the various factors involved and a robust analytical framework. However, by taking a systematic and comprehensive approach to this task, we can gain valuable insights into the development of more robust and reliable biometric systems, informing future research efforts and technology development.

\section{Linear Model of Results}

\begin{table}
    \centering
    \small
    \setlength{\tabcolsep}{4pt}
\begin{tabular}{lcrcr}
\hline
Cov.Name & Value & Coef. & CI & P-val \\
\hline
\hline
Intercept & - & \textbf{-7.003} & (-7.340, -6.666) & 0.000 \\
\hline
Algorithm & \textbf{System A} & +0.00 & - & - \\
Algorithm & System B & +0.27 & (0.21, 0.34) & 0.000 \\
Algorithm & System C & +0.45 & (0.38, 0.51) & 0.000 \\
Algorithm & System D & \textbf{+0.69} & (0.63, 0.76) & 0.000 \\
Algorithm & System E & +0.41 & (0.34, 0.47) & 0.000 \\
\hline
Has Gait & \textbf{False} & +0.00 & - & - \\
Has Gait & True & -0.28 & (-0.33, -0.23) & 0.000 \\
\hline
Has Turb. & \textbf{False} & +0.00 & - & - \\
Has Turb. & True & +0.06 & (-0.01, 0.13) & 0.104 \\
\hline
Head Hgt & \textbf{$>$90 Pix} & +0.00 & - & - \\
Head Hgt & 60-90 Pix & +0.48 & (0.38, 0.58) & 0.000 \\
Head Hgt & 50-60 Pix & \textbf{+0.52} & (0.41, 0.62) & 0.000 \\
Head Hgt & 40-50 Pix & \textbf{+0.67} & (0.58, 0.77) & 0.000 \\
Head Hgt & 30-40 Pix & \textbf{+1.32} & (1.22, 1.43) & 0.000 \\
Head Hgt & $<$30 Pix & \textbf{+1.88} & (1.73, 2.04) & 0.000 \\
Head Hgt & Restricted & \textbf{+2.23} & (2.15, 2.31) & 0.000 \\
\hline
Modality & \textbf{Face} & +0.00 & - & - \\
Modality & Body & \textbf{+0.74} & (0.64, 0.84) & 0.000 \\
\hline
Camera Loc & \textbf{Ctrl} & +0.00 & - & - \\
Camera Loc & Short-Range & -0.34 & (-0.49, -0.20) & 0.000 \\
Camera Loc & Med-Range & \textbf{+0.95} & (0.79, 1.11) & 0.000 \\
Camera Loc & Long-Range & \textbf{+1.95} & (1.72, 2.19) & 0.000 \\
Camera Loc & Elevated & +0.21 & (0.11, 0.32) & 0.000 \\
Camera Loc & Uav & \textbf{+1.00} & (-0.12, 2.12) & 0.081 \\
\hline
Solar Load & \textbf{0-300} & +0.00 & - & - \\
Solar Load & 300-600  & -0.20 & (-0.27, -0.13) & 0.000 \\
Solar Load & 600-900  & \textbf{+0.51} & (0.43, 0.59) & 0.000 \\
Solar Load & Above 900 & \textbf{+0.87} & (0.79, 0.94) & 0.000 \\
\hline
Wind Speed & \textbf{0-3 M/S} & +0.00 & - & - \\
Wind Speed & 3-6 M/S & -0.16 & (-0.21, -0.10) & 0.000 \\
Wind Speed & 6-9 M/S & -0.04 & (-0.14, 0.06) & 0.412 \\
Wind Speed & 9-12 M/S & +0.27 & (0.01, 0.52) & 0.039 \\
\hline
Temperature & \textbf{Below 0 C} & +0.00 & - & - \\
Temperature & 0-10 C & +0.11 & (0.03, 0.20) & 0.007 \\
Temperature & 10-20 C & +0.33 & (0.18, 0.48) & 0.000 \\
Temperature & 20-30 C & -0.34 & (-0.50, -0.18) & 0.000 \\
Temperature & 30-40 C & -0.14 & (-0.33, 0.04) & 0.132 \\
\hline
\end{tabular}
    \caption{This table shows how covariate categories relate to expected model performance.  Coefficients can be used to quickly estimate the impact to FAR rates while the p-values and confidence intervals demonstrate statistical significance of results. }
    \label{tab:linear_mixed_model_coef}
\end{table}

\begin{figure*}
    \centering
    \includegraphics[width=0.65\linewidth]{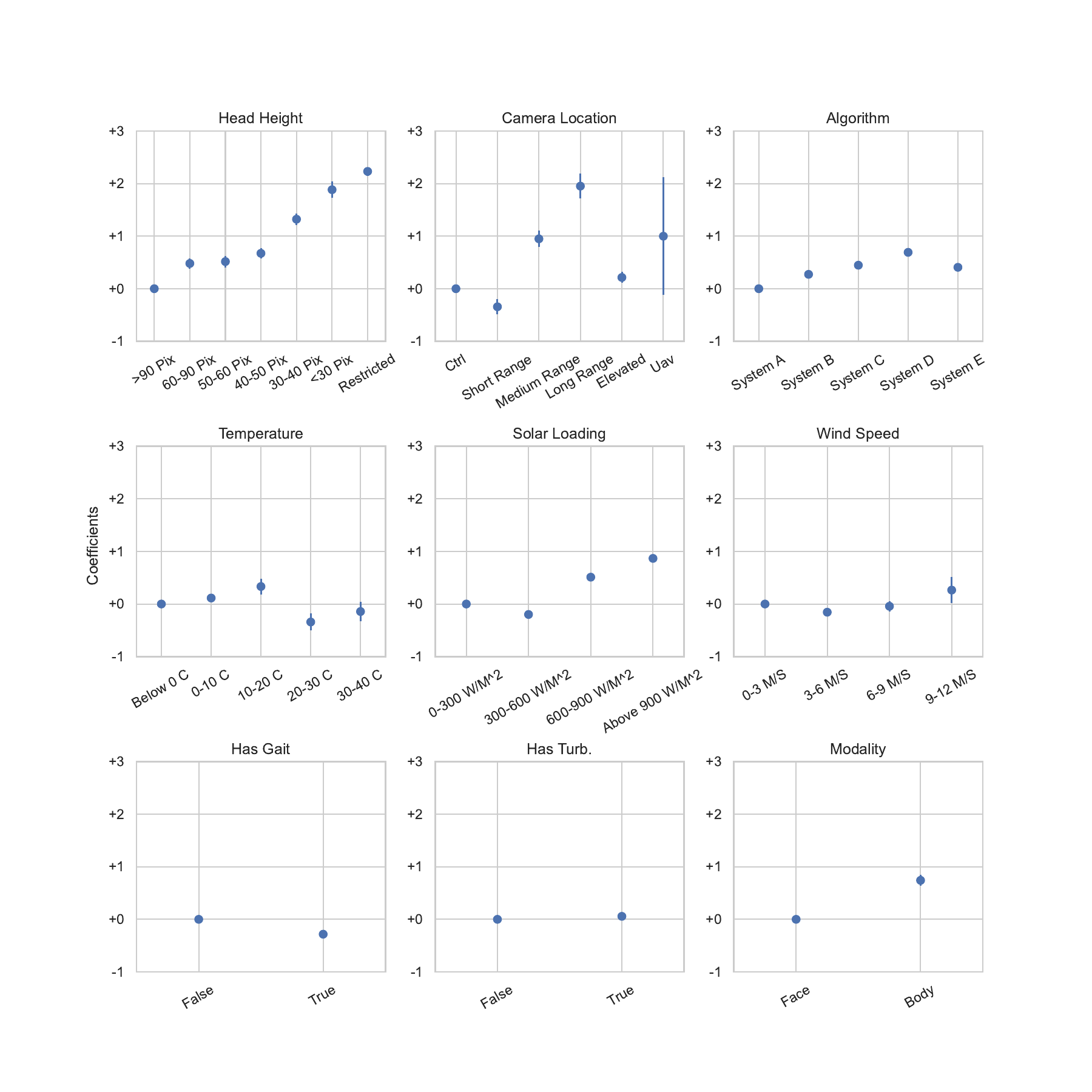}
    \caption{Visual depiction of the linear model results.  Each change of $\pm 1.0$ to the coefficients will result in approximately an order of magnitude change to the expected FAR.  Positive values indicate more difficulty in recognition. Confidence intervals for these estimates are shown to help with understanding the statistical significance of the results.}
    \label{fig:coefficients-visual}
\end{figure*}

To model the performance of the biometric recognition algorithms, we used a linear mixed model approach. This approach allows us to represent the relationship between the covariates and the average genuine acceptance score (in this case, the normalized score) in a mathematically straightforward and interpretable manner. By employing linear mixed models, we can account for the potential correlations between repeated measures of the camera setups and locations, ensuring that our analysis accurately reflects the variability and uncertainty in the data.

This modeling approach at a fundamental level is very simple in that it predicts the average genuine score for a specified set of covariates. Unlike conventional biometric algorithm evaluations which produce an ROC or DET curve, this approach, which may initially seem unconventional, predicts a simple score which is representative of the genuine scores rather than determining a TAR or FAR. 

To interpret the model, it is assumed that the TAR is kept constant at 50\% as discussed earlier, while the FAR is allowed to vary. However, it is possible to easily estimate the FAR due to the score normalization that has been applied. This approach enables a clear understanding of the model's performance in predicting genuine scores and how it may impact the FAR. As will be seen in the following discussion, while holding the TAR rate fixed at 50\%, the model predicts a broad range of selectivity with a base FAR of 1 in 10,000,000 under ideal conditions. When adding challenging conditions like very long distances, low resolution, non-ideal weather, the associated FAR can increase by many orders of magnitude.

The analysis in this work is significantly influenced by the study conducted by Beveridge et al.~\cite{beveridge2010frvt} which applied GLMM to model covariate interactions for face recognition algorithms. Their approach, however, was different in several aspects. Primarily, their focus was on predicting TAR using logistic regression, whereas our model predicts transformed scores, which directly correlate with algorithm performance. Although TAR is commonly used to assess algorithm performance, the transformation we used in our work is linear, offering an intuitive understanding of the model coefficients.

In comparison to the Beveridge model, our approach is simpler, without the need to explore significant interactions and employ model selection techniques to identify the most crucial factors affecting the performance. Although interactions may indeed play a significant role in this research, we have opted for a simpler model. The Beveridge work, although more complex, does provide simple graphics to show the impact of covariate combinations. However, our model can be interpreted by adding the model coefficients.

To keep the models simpler, we converted all covariates into a small number of categorical values. This process groups the covariates into reasonably sized pools, allowing us to represent the impact of each factor in a manageable and interpretable way. By utilizing the linear mixed model approach, and converting the covariates into categorical values, our approach ensures that our statistical model accurately represents the performance of the biometric recognition algorithms and provides a robust basis for identifying the key factors that influence system performance.

During the pre-processing stage, we identified a need to drop certain covariates from the dataset due to instability in the models caused by missing values. Specifically, 900 of the 9,215 probes were dropped from the dataset due to missing weather information, which is a critical factor in predicting performance. Additionally, 20 probes associated with one individual were dropped because the individual was labeled as ``Unspecified'' sex. Although demographics are not the focus of this study, it is an important topic the authors plan to investigate in future work.

As discussed in Section \ref{sec:correlated_variables}, the dataset includes two covariates, collection id and sensor model, which can act as challenging confounding variables. To mitigate their effect on the model, we included these covariates as groups in the mixed model. Given that sensor models are typically distributed for each collection and therefore performance may change and interact between collection to collection for these two variables, it is essential to account for their potential impact on the model results.

By considering all possible combinations of $ (\text{sensor model} \times \text{collection id} )$, we arrived at 63 unique groups in the mixed-effects model. This approach allows us to account for the potential variability and interaction between the sensor models and collection locations, ensuring that our statistical model accurately predicts the performance of the systems and provides a robust basis for isolating the key factors that influence system performance. In terms of implementing this grouping relative to the model, we include these grouped covariates as fixed effects in the linear mixed model. This approach allows us to estimate the average effect of each group on the genuine score, while accounting for the potential correlation with the other covariates of interest. 

The resulting model, presented in Table~\ref{tab:linear_mixed_model_coef}, showcases the impact of each covariate category on the mean score with most covariates deemed statistically significant within a confidence interval of $\pm 0.11$. Note that the UAV category is a notable outlier with a much larger confidence interval due to the variety of quality related issues and a somewhat smaller number of data points in that category.  In the table, we have chosen to bold the results with a coefficient of 0.5 or higher to call attention to the most impactful results. The same results are shown visually in Figure~\ref{fig:coefficients-visual}

This analysis enables the estimation of a specific point on the ROC curve. By referencing the coefficients provided in the table, one can estimate the  genuine score. The variation in the FAR can then be estimated using the following equation, where $S$ represents the genuine score estimate by the model:

\begin{equation}
\text{FAR}_{est} = 10^{S}
\end{equation}

This analysis selects default categories for the model, which are typically the easiest cases, such as close-range, eye-level cameras, and a head size greater than 90 pixels. Although these scenarios are relatively straightforward, the data collection process in this analysis is still challenging due to the outdoor environment and \textbf{uncooperative subjects} scenario. The intercept in the model represents this default case as depicted with bold text for each covariate value in Table~\ref{tab:linear_mixed_model_coef}. By assuming a TAR of 0.5, the FAR can be estimated for this easy scenario that conveniently works out to approximately a 1 in 10,000,000 FAR:

 \begin{equation}
 \text{FAR}_{est} = 10^{-7.003} = \frac{1}{10,023,052}
 \end{equation}

Changing the value of the covariates of interest will typically make recognition more difficult than this baseline.  For example, we can very quickly estimate that a camera configuration where the \text{head height} is ``$< 30$ pixels'' (+1.884) and the camera is located at long-range (+1.952) will increase the FAR error by 4 orders of magnitude while the TAR estimate is assumed to be constant at 0.5.

 \begin{equation}
 \text{FAR}_{est} = 10^{-7.003+1.884+1.952} = \frac{1}{1,469}
 \end{equation}

This shows that the FAR can change by 4 orders of magnitude by just changing 2 important covariates.

\section{Discussion}

As mentioned in Section \ref{sec:correlated_variables}, one of the main factors driving performance is the choice of sensor model/camera which encapsulates the hardware, optics, and software and to some extent the camera configuration. For example, in the data collection the cameras are always setup to capture images with the best possible visual quality and with the highest quality compression to maintain as much detail and accuracy is retained from the original video. However, due to the variety of cameras and complexity of modeling the whole camera performance, this selection is represented in our model as a fixed effect and therefore is not shown in the final table of coefficients.

Two related covariates that are two of the strongest predictors of accuracy are \textbf{head height} in pixels, and \textbf{modality}.  It should be mentioned that these likely are correlated since \textbf{modality} relates to how the zoom level of the camera is set.  For ``face'' configuration, the camera is zoomed-in to focus on the upper body to capture more fine details.  For ``body'', the camera is zoomed out to capture the full length of the body as well as a full 10-meter long walking area to support gait recognition.  The model results show that the ``face'' configuration is strongly preferred which will typically be associated with head heights with more pixels.

The \textbf{head height} is certainly one of the strongest predictors of accuracy and has a very straight forward interpretation: more pixels on target makes recognition easier.  This covariate also has a special and very difficult category which is ``face restricted'' where the face is either too small or the subject is not facing the camera, in which case the systems cannot rely on the face recognition modality.

The \textbf{camera location} covariate is probably of the most interest for the research program. On the ground, it would seem there is only a minimal difference between ``ctrl'' (close-range) and ``short-range'' locations meaning that with a good setup and recognition systems in place, recognition at extended distances is possible. At ``medium'' ($250m-550m$) to ``long'' ($>550m - 1000m$) range the problem gets much more difficult with FAR increasing by almost two orders of magnitude.

The elevated locations show a similar story with only a minor difference between the close-range cameras at ground level, ``ctrl'', and the "elevated" cameras on masts, which are all colocated with the ``ctrl'' cameras on the group. ``UAV'' mounted cameras exhibit significant variability. This likely results from the difference between small, modern HD and 4k quadcopters that typically fly at low altitude and within 100m of the subjects and larger systems that fly at longer distances and up to 1200 ft altitude, often with older or lower resolution payloads. In this model, all of these are grouped into the same category.

Weather results are also interesting with most weather related effects showing only modest impact on accuracy.  \textbf{Solar loading} had the most effect with ``$ > 900 \frac{\textbf{W}}{\textbf{m}^2} $'' increasing the FAR by almost an order of magnitude.  This should correspond to direct sunlight conditions that have been known to cause issues with outdoor imaging when compared to the smooth lighting available from overcast skies.
 
\textbf{Turbulence} and its mitigation represent a significant challenge in addressing the long-range biometrics problem and is an important research topic for the program. Despite this, the impact of turbulence on the recognition systems was found to be minimal, with no statistically significant effect observed. The systems may be mitigating the impact of turbulence by integrating through time or capturing ``lucky'' frames \cite{10449024}. It is also possible that turbulence is already accounted for in the model by other covariates, such as \textbf{distance} or \textbf{solar loading}. However, it should be noted that high turbulence data is only an issue for a limited portion of the test data. \textbf{Turbulence} only really affects the image quality for medium- and long-range cameras and of those, we could only measure turbulence ($CN^2$) for a small number of cameras which were placed near collection scintillometers. Consequently, estimating turbulence for the majority of the cameras is infeasible.

\section{Future Work}
As this is the first paper that looks at the effects of covariates on fused whole body recognition problem, there is still plenty of interesting topics that can be explored in the future.  The BRIAR datasets used in this analysis represent a complex set of covariates and there are many topics that have not yet been explored including demographics, clothing features, yaw and pitch angles, video compression, etc.  

Also of interest is the algorithm modality.  In this study, we only look at the fusion of all features that these systems extract from the video of the subjects; however, internally the systems are composed of face, body, and gait specific components.  While numerous studies have been conducted on face performance, the body and gait based features are relatively new and may yield interesting differences between biometric modalities.

Finally, the model presented here has been intentionally kept simple.  A small number of covariates were examined since they were known to influence performance and because they are of high interest to the research program as a whole. The simplicity of the model allows for easy interpretation, however there are some valuable additions that could be adopted from the work in \cite{beveridge2010frvt}. First, model selection was used to reduce the model to a minimum number of covariates that effect performance, and secondly, that model explored covariate interactions.  While these changes may result in a more complex analysis, it may also allow us to include more covariates and better understand how these variables influence system performance.

\section{Conclusions}

In this study, we have investigated the relationship between various covariates and the performance of fused whole body biometric recognition algorithms, specifically focusing on biometric recognition at altitudes, ranges, and elevated camera positions in complex and dynamic environments. By analyzing sensor models and collection locations as fixed effects, we account for these factors as confounders. The analysis provides insights for operational systems facing a wide range of challenges, including close-range and long-range recognition, elevated cameras, weather conditions, and other factors that can impact algorithm performance. Understanding these factors is crucial for the advancement of cutting-edge biometric technologies and supporting the development of more robust and reliable detection methods in challenging environments.

By taking a systematic approach with this dataset, we have gained significant insights into the challenges and interactions that must be considered when designing new systems to function in complex and dynamic environments. This comprehensive analysis will serve as a foundation for further understanding of biometric recognition system performance in various conditions and provide valuable insights guiding future research and development efforts related to identifying people in this complex scenario.

\section*{Acknowledgments}

This research is based upon work supported by the Office of the Director of National Intelligence (ODNI), Intelligence Advanced Research Projects Activity (IARPA), via D20202007300010. The views and conclusions contained herein are those of the authors and should not be interpreted as necessarily representing the official policies, either expressed orimplied, of ODNI, IARPA, or the U.S. Government. The U.S. Government is authorized to reproduce and distribute reprints for governmental purposes notwithstanding any copyright annotation therein.

This research used resources from the Knowledge Discovery Infrastructure at the Oak Ridge National Laboratory, which is supported by the Office of Science of the U.S. Department of Energy under Contract No. DE-AC05-00OR22725

Notice:  This manuscript has been authored by UT-Battelle, LLC, under contract DE-AC05-00OR22725 with the US Department of Energy (DOE). The US government retains and the publisher, by accepting the article for publication, acknowledges that the US government retains a nonexclusive, paid-up, irrevocable, worldwide license to publish or reproduce the published form of this manuscript, or allow others to do so, for US government purposes. DOE will provide public access to these results of federally sponsored research in accordance with the DOE Public Access Plan
(\url{http://energy.gov/downloads/doe-public-access-plan}).

{\small
\bibliographystyle{unsrt}
\bibliography{egbib}

\begin{thebibliography}{10}

\bibitem{nalty2022brief}
Chrisopher~B Nalty, Neehar Peri, Joshua Gleason, Carlos~D Castillo, Shuowen Hu,
  Thirimachos Bourlai, and Rama Chellappa.
\newblock A brief survey on person recognition at a distance.
\newblock In {\em 2022 56th Asilomar Conference on Signals, Systems, and
  Computers}, pages 145--152. IEEE, 2022.

\bibitem{liu2024farsight}
Feng Liu, Ryan Ashbaugh, Nicholas Chimitt, Najmul Hassan, Ali Hassani, Ajay
  Jaiswal, Minchul Kim, Zhiyuan Mao, Christopher Perry, Zhiyuan Ren, et~al.
\newblock Farsight: A physics-driven whole-body biometric system at large
  distance and altitude.
\newblock In {\em Proceedings of the IEEE/CVF Winter Conference on Applications
  of Computer Vision}, pages 6227--6236, 2024.

\bibitem{Zhu_2024_WACV}
Haidong Zhu, Wanrong Zheng, Zhaoheng Zheng, and Ram Nevatia.
\newblock Sharc: Shape and appearance recognition for person identification
  in-the-wild.
\newblock In {\em Proceedings of the IEEE/CVF Winter Conference on Applications
  of Computer Vision (WACV)}, pages 6290--6300, January 2024.

\bibitem{nikhal2023weakly}
Kshitij Nikhal and Benjamin~S Riggan.
\newblock Weakly supervised face and whole body recognition in turbulent
  environments.
\newblock In {\em 2023 IEEE International Joint Conference on Biometrics
  (IJCB)}, pages 1--10. IEEE, 2023.

\bibitem{10449024}
Chun~Pong Lau, Maitreya Suin, and Rama Chellappa.
\newblock Atdetect: Face detection and keypoint extraction at range and
  altitude.
\newblock In {\em 2023 IEEE International Joint Conference on Biometrics
  (IJCB)}, pages 1--10, 2023.

\bibitem{myers2023recognizing}
Blake~A Myers, Lucas Jaggernauth, Thomas~M Metz, Matthew~Q Hill, Veda~Nandan
  Gandi, Carlos~D Castillo, and Alice~J O’Toole.
\newblock Recognizing people by body shape using deep networks of images and
  words.
\newblock In {\em 2023 IEEE International Joint Conference on Biometrics
  (IJCB)}, pages 1--8. IEEE, 2023.

\bibitem{huang2023whole}
Siyuan Huang, Ram~Prabhakar Kathirvel, Chun~Pong Lau, and Rama Chellappa.
\newblock Whole-body detection, recognition and identification at altitude and
  range.
\newblock {\em arXiv preprint arXiv:2311.05725}, 2023.

\bibitem{Azad_2024_CVPR}
Shehreen Azad and Yogesh~Singh Rawat.
\newblock Activity-biometrics: Person identification from daily activities.
\newblock In {\em Proceedings of the IEEE/CVF Conference on Computer Vision and
  Pattern Recognition (CVPR)}, pages 287--296, June 2024.

\bibitem{doers2023}
Dawei Du, Cole Hill, Gabriel Bertocco, Mauricio~Pamplona Segundo, Wes Robbins,
  Brandon RichardWebster, Roderic Collins, Sudeep Sarkar, Terrance Boult, and
  Scott McCloskey.
\newblock Doers: Distant observation enhancement and recognition system.
\newblock In {\em 2023 IEEE International Joint Conference on Biometrics
  (IJCB)}, pages 1--11, 2023.

\bibitem{cornett2023expanding}
David Cornett, Joel Brogan, Nell Barber, Deniz Aykac, Seth Baird, Nicholas
  Burchfield, Carl Dukes, Andrew Duncan, Regina Ferrell, Jim Goddard, et~al.
\newblock Expanding accurate person recognition to new altitudes and ranges:
  The briar dataset.
\newblock In {\em Proceedings of the IEEE/CVF Winter Conference on Applications
  of Computer Vision}, pages 593--602, 2023.

\bibitem{beveridge2010frvt}
J~Ross Beveridge, Geof~H Givens, P~Jonathon Phillips, Bruce~A Draper, David~S
  Bolme, and Yui~Man Lui.
\newblock Frvt 2006: Quo vadis face quality.
\newblock {\em image and Vision Computing}, 28(5):732--743, 2010.

\end{thebibliography}
}
\end{document}